\documentclass[journal]{IEEEtran}

\ifCLASSINFOpdf
\else
\fi
\usepackage{amsmath,graphicx,algorithm,algorithmic,tabularx,subfigure}
\usepackage{multirow}
\usepackage{mathrsfs}
\usepackage{hyperref}

\begin{document}

\title{An Iteratively Optimized Patch Label Inference Network for Automatic Pavement Distress Detection}
%
%
%

\author{Wenhao Tang,
        Sheng Huang,~\IEEEmembership{Member,~IEEE,}
        Qiming Zhao,
        Ren Li,
        and~Luwen Huangfu
\thanks{S. Huang is with Ministry of Education Key Laboratory of Dependable Service Computing in Cyber Physical Society, Chongqing, 400044, P.R.China,
W. Tang, Q. Zhao and S. Huang are with the School of Big Data and Software Engineering, Chongqing University, Chonqqing, 400044 P.R.China,(email:\{whtang, qmzhao, huangsheng\}@cqu.edu.cn), S. Huang is the corresponding author.}
\thanks{R. Li is with Chongqing Jiaotong University, 400074, P.R.China, (email:renli@cqjtu.edu.cn)}
\thanks{L. Huangfu is with Fowler College of Business, San Diego State University, San Diego, CA 92182, USA, (email: lhuangfu@sdsu.edu)}}

%
%

\markboth{}%
{Shell \MakeLowercase{\textit{et al.}}: Bare Demo of IEEEtran.cls for IEEE Journals}

\maketitle

\begin{abstract}
We present a novel deep learning framework named the Iteratively Optimized Patch Label Inference Network (IOPLIN) for automatically detecting various pavement distresses that are not solely limited to specific ones, such as cracks and potholes. IOPLIN can be iteratively trained with only the image label via the Expectation-Maximization Inspired Patch Label Distillation (EMIPLD) strategy, and accomplish this task well by inferring the labels of patches from the pavement images. IOPLIN enjoys many desirable properties over the state-of-the-art single branch CNN models such as GoogLeNet and EfficientNet. It is able to handle images in different resolutions, and sufficiently utilize image information particularly for the high-resolution ones, since IOPLIN extracts the visual features from unrevised image patches instead of the resized entire image. Moreover, it can roughly localize the pavement distress without using any prior localization information in the training phase. In order to better evaluate the effectiveness of our method in practice, we construct a large-scale Bituminous Pavement Disease Detection dataset named CQU-BPDD consisting of 60,059 high-resolution pavement images, which are acquired from different areas at different times. Extensive results on this dataset demonstrate the superiority of IOPLIN over the state-of-the-art image classification approaches in automatic pavement distress detection. The source codes of IOPLIN are released on \url{https://github.com/DearCaat/ioplin}, and the CQU-BPDD dataset is able to be accessed on \url{https://dearcaat.github.io/CQU-BPDD/}.
\end{abstract}


\IEEEpeerreviewmaketitle

\section{Introduction}
\IEEEPARstart{P}{avement} distresses pose a great threat to the driving safety of vehicles, since roads age over time due to wear and tear. These distresses reduce the stability of the road surface and form defects of various shapes. Detecting pavement distresses is one of the most important steps for maintaining road stability. The traditional pavement distress detection scheme is mainly manual detection, which requires a large number of professionals and fruitful domain knowledge. Moreover, professional testing requires expensive professional sensors~\cite{andrea2014}. As the total mileage and the usage frequency of roads increase, it is almost impossible to accomplish such a detection task manually. Thanks to the rapid progress in Artificial Intelligence (AI), recent computer vision techniques have been able to provide an elegant and effective way of automatically detecting pavement distresses.

\begin{figure}[t]
\centering
\includegraphics[width=8.5cm]{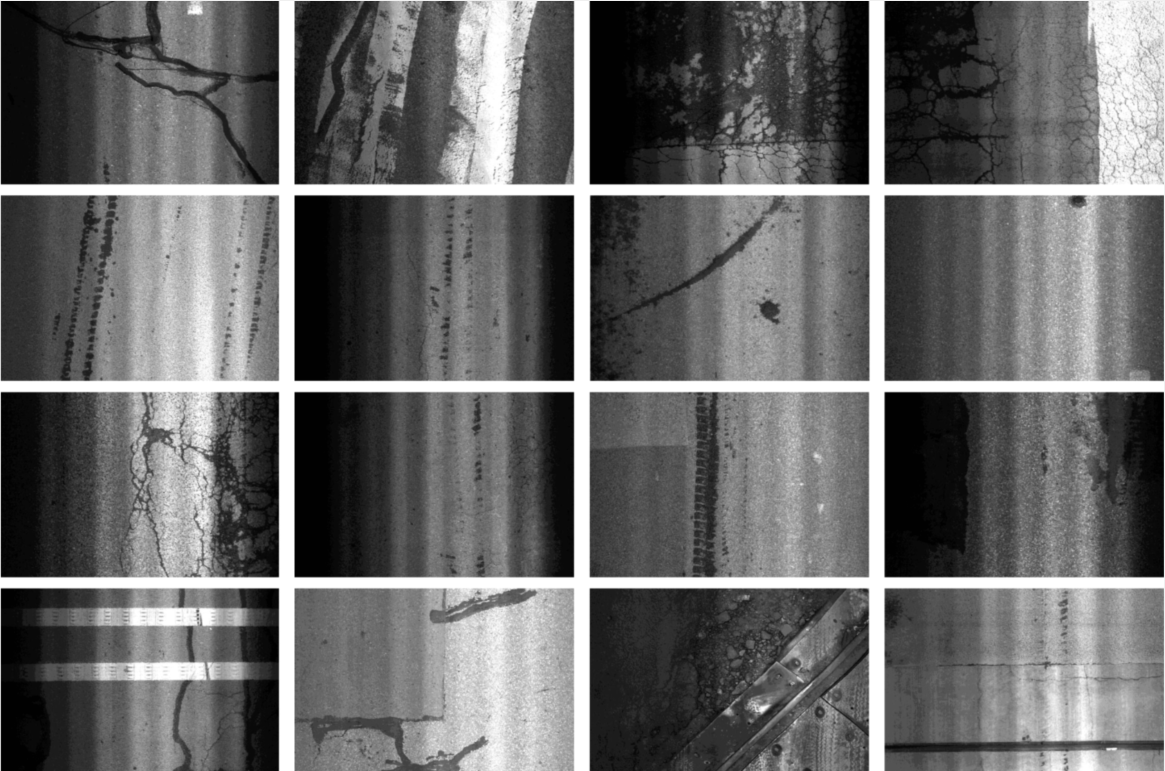}
\vspace{-0.1cm}
\caption{Some examples in CQU-BPDD dataset, which involve various pavement distresses such as alligator crack, raveling, pothole, repair.}
\vspace{-0.5cm}
\label{disimg}
%
\end{figure}
Over decades, there are numerous impressive studies that address different pavement disease analysis issues from the perspective of computer vision. However, the existing pavement distress detection approaches mainly focus on detecting specific pavement diseases, mostly cracks, raveling or potholes~\cite{uav,cnnta,ddpdd,esdl,aprs,siri,qsvm}. Moreover, many so-called pavement distress detection works also ambiguously define the tasks. Most of them actually should be grouped into the pavement crack segmentation~\cite{shi2016,gaborfilter,zhang2016,fphbn,msif} and localization~\cite{mandal2018,wang2018,ale2018}, which are totally different to the real pavement distress detection.
In this paper, the pavement distress detection task we concentrate on is relevant but also quite different from these tasks. We intend to judge whether there are diseases or not based on the pavement image. The pavement distresses that we intend to detect are not only limited to cracks, raveling and potholes but also other general distresses, such as repair and crack pouring. We name this task automatic pavement distress detection which can be deemed as a generalization of these specific distress detection tasks. This task is also an important pre-step of pavement distress segmentation and a core step of pavement distress localization. Although such a task can be considered as a typical pavement image binary classification problem, it is very challenging, since the pavement imaging suffers from uneven illumination, chromatic aberration, road markings in the background, and high diversity in the appearance of various distresses, such as cracks, potholes, erosive pits, and their mixtures, as shown in Figure~\ref{disimg}.

%
%
%


The Convolutional Neural Networks (CNNs) such as ResNet~\cite{he2016} and GoogLeNet~\cite{szegedy2016} are the dominant methods for image classification. In this paper, we also intend to leverage CNN for addressing the automatic pavement distress detection issue. However, these CNN models often translate the image into a fixed low resolution image and then accomplish the classification based on the entire image. Such image translation will lose a lot of image information particularly for the high resolution images. For example, the input of ResNet is fixed to $224\times224$ while the resolution of our pavement image is $1200\times900$. After the image translation, the input image will lose 95\% of pixels. Moreover, the diseased area is often just a very small fraction of the entire pavement image. In such a manner, such aforementioned global-based approaches may be more easily obstructed by noise and background variations. Therefore, we propose a novel local-based deep learning framework named Iteratively Optimized Patch Label Inference Networks (IOPLIN) to address automatic pavement distress detection issues.

In IOPLIN, the pavement image is segmented into dozens of patches, and then an EfficientNet~\cite{tan2019} is considered as a Patch Label Inference Network (PLIN) for inferring the labels of patches. Finally, the detection result of a pavement image will be achieved by the maximum pooling of its inferred patch labels. The main obstacle of this methodology is that only image-level labels are available. To address this issue, we propose the Expectation-Maximization Inspired Patch Label Distillation (EMIPLD) strategy for iteratively and gradually optimizing PLIN only based on the image label. Different from the convolutional CNN-based pavement distress detection regime, IOPLIN not only offers improved detection results in image level, but also roughly localizes the disease in the pavement image via EMIPLD in a weakly supervised manner. To evaluate the effectiveness of our work, we introduce a novel large-scale Bituminous Pavement Disease Detection database named CQU-BPDD consisting of 60,059 high-resolution pavement images that involve seven different diseases and a normal pavement. These images are automatically captured by in-vehicle cameras from different areas in southern China. The extensive experimental results on this dataset validate the effectiveness and superiority of IOPLIN in comparison with state-of-the-art CNN algorithms.

The main contributions are summarized as follows:
\begin{itemize}

\item To the best of our knowledge, we are the first work to formally define and systematically investigate the automatic pavement distress detection task that is not just limited to specific diseases such as cracks, raveling and potholes.

\item We release a novel, large-scale automatic pavement distress detection dataset that is acquired from real scenarios and involves various diseases. All the existing pavement distress datasets only contain hundreds or thousands of samples, while our dataset contains more than sixty thousand high-resolution pavement images and involves more types of diseases.

\item We present a novel, deep learning-based automatic pavement distress detection approach named Iteratively Optimized Patch Label Inference Networks (IOPLIN), which can not only sufficiently utilize the information of any resolution image for detecting pavement distresses but also can roughly localize distress positions solely based on the image label.

\item We conduct extensive experiments to systematically and empirically compare the performances of the recent state-of-the-art CNN approaches in automatic pavement distress detection, and validate the prominent superiority of our work over them not only in performance but also in robustness and cross-data generalization.

\end{itemize}

\section{Related Work}
\subsection{Image-based Pavement Distress Analysis}
The conventional pavement distress analysis approaches are mainly based on low-level image analysis, hand-crafted features and classical classifiers~\cite{morcrack,ndh,gradientfilter,lbp,aprs,siri}. For example, Shi et al.~\cite{shi2016} presented a random structured forest named CrackForest, that was combined with the integral channel features for automatic road crack detection. In~\cite{gaborfilter}, a filter bank consisting of multiple oriented Gabor filters is proposed to detect road cracks. Pan et al.~\cite{uav} leveraged KNN, SVM, random forest and neural networks to recognize the pavement cracks and potholes based on the images acquired by Unmanned Aerial Vehicle (UAV). Hadjidemetriou et al.~\cite{qsvm} leveraged the traditional support vector machine to detect the pavement patch. Nhat-Duc Hoang~\cite{sgdlr} utilizes using image texture based feature extraction and stochastic gradient descent logistic regression for automatic detection of asphalt pavement raveling.

Inspired by the recent remarkable successes of deep learning in extensive applications, there has been a trend of more and more researchers applying the advanced deep learning approaches to tackle these tasks~\cite{spcnn,cnnta,ddpdd,regioncnn}. Zhang et al.~\cite{zhang2016} segmented the pavement cracks by detecting the crack point with Convolutional Neural Networks (CNN). In~\cite{ddpdd}, an ImageNet pre-trained VGG-16 DCNN is applied to categorize the pavement image into "crack" or "non-crack". Xia~\cite{xia2018} adopted the Single Shot multibox Detector~(SSD)~\cite{liu2016} networks for localizing the pavement disease. Some researchers also utilize some well-known object detection frameworks, such as YOLO v2, Faster RCNN and RetinaNet, to localize the pavement diseases~~\cite{mandal2018,wang2018,ale2018,pavementYOLO}. Fan et al.~\cite{cnnta} produced a novel automatic road crack detection system. In this system, a CNN was used for determining whether the pavement image contains cracks or not, and then an adaptive thresholding method was presented for segmenting the cracks based on the image smoothed by bilateral filters.

In summary, the tasks of the aforementioned works can be grouped into three categories: pavement crack segmentation~\cite{shi2016,gaborfilter,zhang2016,fphbn,msif}, pavement crack localization~\cite{mandal2018,wang2018,ale2018} and specific pavement distress detection~\cite{uav,cnnta,ddpdd,aprs,siri,qsvm}, such as crack and raveling detections. Clearly, the general pavement distress detection, which is not just limited to detect specific diseases, still remains unstudied systematically. Moreover, most of the existing pavement distress datasets, such as Crack Forest Dataset (CFD)~\cite{shi2016}, CrackTree200~\cite{cracktree}, Crack500~\cite{fphbn}, are mainly designed for pavement crack segmentation. These datasets generally contain hundreds or thousands of samples and only the diseased images are involved, which cannot be directly applied for studying pavement distress detection.

In this paper, we attempt to develop a novel deep learning method named Iteratively Optimized Patch Label Inference Network (IOPLIN) for automatically detecting pavement distresses not just limited to some specific ones. And a large-scale pavement disease image dataset named CQU-BPDD is released for supporting the study of automatic pavement distress detection. The CQU-BPDD dataset is a more challenging pavement image dataset, which contains 60,059 high-resolution pavement images involving seven different types of pavement diseases and also the normal case.

\subsection{Object Detection and Image Classification}
Object detection, as one of the most fundamental and challenging problems in computer vision, has received extensive attention in recent decades. The goal of object detection is to determine whether there are any instances of objects from given categories, such as humans, cars, bicycles, dogs or cats, in an image and if present, to return the spatial location and extent of each object instance~\cite{godsurvey,objectsurvey}. The traditional detectors often utilize the sliding window method to collect a set of object proposals or candidate boxes from an image, and then an elaborated handcraft feature representation method is applied for representing each object proposal. Finally, the object detection task is deemed as the object proposal binary classification problem for solution. The core of these approaches is the proposal handcraft representation. The representative representations include Haar wavelet feature~\cite{vj},  Histogram of Oriented Gradient (HOG)~\cite{hog},  Local Binary Pattern (LBP)~\cite{lbp,hoglbp} and so on. Moreover, the learning-based representation methods are also the popular way for object detection. The representative method of this category is Fisher-Vector (FV). In many object detection tasks, FV and its variants show better performances over the handcraft features~\cite{fisher,sdod}.

In the last decade, deep learning approaches have become the mainstream methodology for object detection, since deep convolutional networks are able to learn robust and high-level feature representations of an image. The deep learning-based object detection approaches can be further grouped into two categories by learning scheme: two-stage detector and one-stage detector. The well-known two-stage detectors include RCNN~\cite{rcnn}, SPPNet~\cite{sppnet}, Fast RCNN~\cite{fastrcnn}, Faster RCNN~\cite{ren2015} and Feature Pyramid Networks~\cite{fpn}. Different from the two-stage detectors, the one-stage detector optimizes the object proposal selection and classification jointly in an end-to-end learning manner. YOLO~\cite{redmon2016}, Single Shot MultiBox Detector (SSD)~\cite{liu2016} and RentiaNet~\cite{lin2017} are the representative one-shot detectors. Since those aforementioned deep learning-based approaches are often beyond the sliding window method for object proposal selection, the accurate location annotation of an object should be provided during the training phase in these approaches.

The setting of our task is very similar to the conventional sliding window-based object detection fashion that the candidate boxes are cropped from the image, and then the detection task is simply degraded into a binary image classification task, since each pavement image of CQU-BPDD dataset is collected by the camera in a professional pavement inspection vehicle which is corresponding to a $2\times3$ meters pavement patch, and can be deemed as a pavement distress detection candidate box of a road. Therefore, the core of the pavement distress detection task is image classification. Currently, the most dominant image classification approach is the CNN-based method. In the recent decade, many classical CNN models have been proposed, such as AlexNet~\cite{alexnet}, VGGNet~\cite{simonyan2014},  GoogleNet~\cite{szegedy2016}, ResNet~\cite{he2016}, EfficientNet~\cite{tan2019}, MobileNet~\cite{mobilenet} and so on. The main problem for applying these advanced methods to automatic pavement distress detection is that the input image should be resized to the low-resolution one for meeting the requirement of the input size of CNNs. Since the pavement image captured by the professional inspection vehicle is high-resolution, it is unavoidable to lost the most visual information of a pavement image during the CNN training. In such a manner, we develop a novel CNN-based image classification method to extract the features from the image patch instead of the entire image in this paper.

\section{METHODOLOGY}
\label{sec:majhead}

\begin{figure*}[ht]
\centering
\includegraphics[width=17cm]{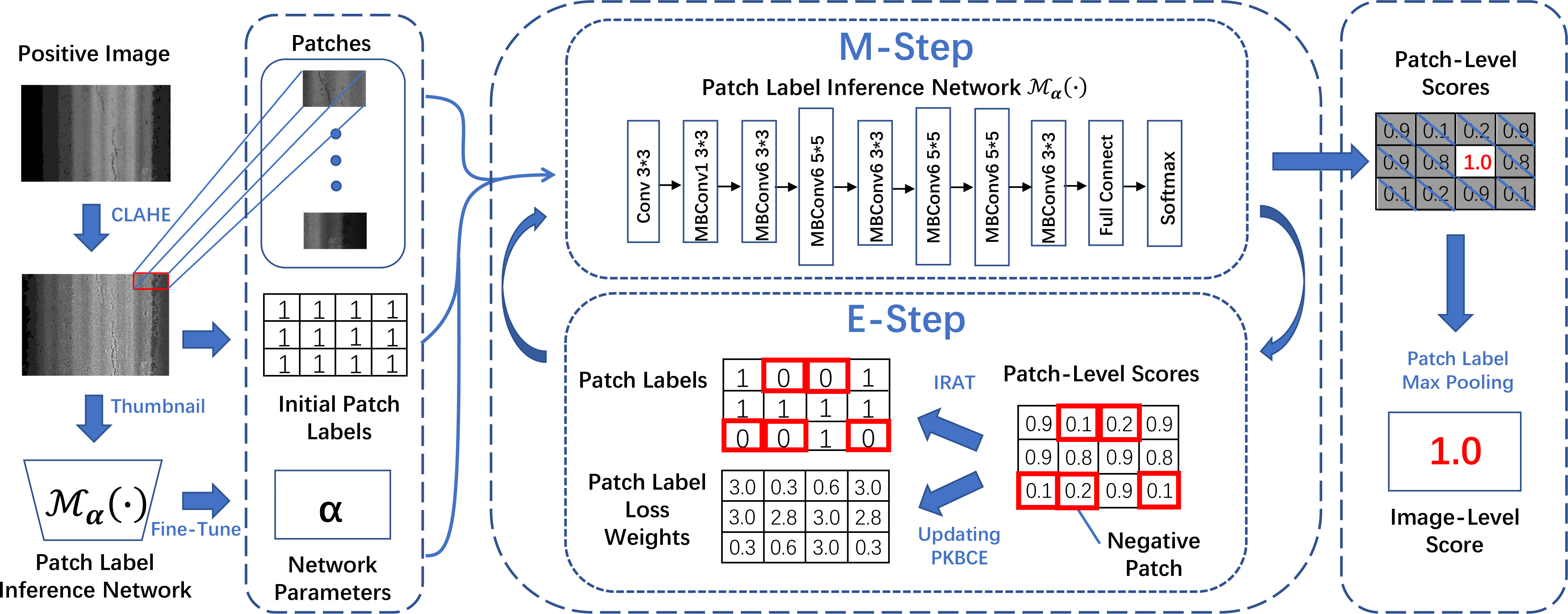}
\vspace{-0.2cm}
\caption{The overview of our method.}
\label{fig:res}
\vspace{-0.4cm}
\end{figure*}

\subsection{Problem Formulation and Overview}
\label{ssec:subhead}

Let $I_i$ be a pavement image associated with a binary label $y_i$ which indicates whether diseases exist or not. The automatic pavement disease detection is essentially a binary image classification task that aims to derive a detector $\mathcal{D}:I_i\to y_i$ to classify a pavement image into "diseased" or "normal".

To tackle the automatic pavement disease detection task, we present a novel, deep learning approach named Iteratively Optimized Patch Label Inference Networks (IOPLIN). In IOPLIN, pavement images are preprocessed by Contrast Limited Adaptive Histogram Equalization (CLAHE)~\cite{pizer1990} to suppress the negative effect of uneven illumination first. The processed image will be further segmented into patches and a Patch Label Inference Networks (PLIN) is trained to infer the patch labels. Finally, the pavement image label can be obtained by the maximum pooling of its patch labels. The core of our approach is the PLIN. However the PLIN cannot be well trained directly, since only the image label is available, whereas the patch labels of each image are unavailable in the training phase. To overcome this difficulty, we present the Expectation-Maximization Inspired Patch Label Distillation (EMIPLD) strategy for iteratively optimizing PLIN training by reasonably initializing patch labels. In the next subsections, we will go into the details of our method.

\subsection{Histogram Equalization and Patches Collection}
\label{he}
Since the pavement images are captured at different times and from different areas, they suffer from serious and uneven illuminations. To suppress the negative impacts of illumination, the pavement image is processed by CLAHE \cite{pizer1990}. The empirical analysis also implies that such preprocessing indeed improves detection performance.

The traditional Convolutional Neural Networks (CNN), such as VGGNet~\cite{simonyan2014}, GoogLeNet~\cite{szegedy2016}, and ResNet~\cite{he2016}, often require input image size around 300$\times$300 while the size of the pavement image on our dataset is 1200$\times$900. Instead of resizing the high-resolution image into the low-resolution one and directly inputting it into the CNN for yielding the final detection results, our approach aims to partition the image into patches and perform the detection by inferring the patch labels using CNN. In such a manner, the image information can be fully exploited, and the side products such as patch labels or patch-based disease confidences can be produced. That may offer a good explanation of the results or benefit the solutions of other follow-up tasks.

In our case, we simply follow the non-overlapping image blocking strategy and fix the patch size to 300$\times$300, since the size of our backbone network (EfficientNet-B3) input is 300$\times$300 and our 1200$\times$900 resolution pavement images can be evenly divided in such a manner. With regard to other resolution pavement images, we can empirically design the image block strategy and the patch size based on the type of backbone network and the size of the pavement image which all pixels of the pavement image are expected to be evenly exploited.

We assume each image is partitioned into $m$ 300$\times$300 patches. Such step can be mathematically denoted as follows:
\begin{equation}\label{}
x_{i} = \mathcal{H}(I_i):=\left \{ p^i_{1},\cdots,p^i_{t},\cdots,p^i_{m} \right \},
\end{equation}
where $x_i$ is the $i$-th pre-processed image, $\mathcal{H}(\cdot)$ is the CLAHE operation and $p^i_t$ represents the $t$-th patch of image. $m$ is the number of patches and equals 12 in our implementation. We also assume there are $n$ images for training. As a result, the total number of patches for training is $n\times m$.

\subsection{Patch Label Inference Network}
\label{plin}
There are many classical CNN models that have proved their effectiveness in image classification. We empirically evaluate several CNN models that have a similar size in parameters and eventually choose the very recent CNN model named EfficientNet-B3 as our backbone network for inferring the labels of patches. This network is pre-trained with the ImageNet dataset and its output layer is replaced with a two-node output layer. For details of EfficientNet-B3, please refer to~\cite{tan2019}. We name this network Patch Label Inference Network (PLIN), and the patch label inference is denoted as follows:
\begin{equation}\label{network}
g^i_t=\mathcal{M}_\alpha(p^i_t),
\end{equation}
where $\mathcal{M}(\cdot)$ is the mapping function of PLIN and $\alpha$ is its associated network parameters. $g^i_t \in (0,1)$ is the prediction value of the true patch label $l^i_t$ where its value is equal to 1 or 0 when there exists or does not exist disease .

\subsection{EM-Inspired Patch Label Distillation}
\label{ssec:subhead}
Unfortunately, only the image label $y_i$ is available while the ground truth of patch label $l^i_t$ is unavailable, which impedes the normal training of PLIN. In this section, we will introduce an iteratively PLIN training strategy named Expectation-Maximization Inspired Patch Label Distillation (EMIPLD). The basic idea of EMIPLD is to give a reasonable initialization of patch label for training a PLIN as well as retrain the PLIN based on the new labels inferred by the previous version PLIN. These steps are iteratively executed until convergence. Considering the training step as $M$ step and the label inference step is $E$ step, such an iteration scheme is very similar to the idea of Expectation-Maximization (EM) algorithm, and the patch labels will be progressively refined during the iteration, just as its name says. Such an idea can work, since the labels of the patches from the normal pavement images are always normal, and consequently these credibly labeled data drive the continual optimization of PLIN and the progressive distillation of the patches from the diseased pavement image.

\subsubsection{Initialization of Patch Labels}
We consider the image label as the initial label of its patch $\tilde{l}^i_{t(0)}=y_i \subset \{0,1\}$. In such a case, the labels of the patches from the normal pavement images are credible while the ones from the diseased pavement images are suspicious, since the diseased areas may not cover all the images.

\subsubsection{The Maximization ($M$) Step}
\label{sssec:subsubhead}

We train PLIN with all training data $X:=\{x_i|\forall i\}$ and their associated current patch labels $L_{(j)}:=\{\tilde{l}^i_{t(j)}|\forall i , t\}$ to achieve the network parameters of PLIN $\alpha_j$ in the $j$-th iteration.

\subsubsection{The Expectation ($E$) Step}
\label{sssec:subsubhead}

The $E$ step is to leverage the trained PLIN to infer the labels of patches. According to Equation~\ref{network}, each patch can achieve a label prediction value referred to as the confidence score $g^i_{t(j)}$. We present the Image-based Rank Aware Threshold (IRAT) scheme for adaptively updating the label of each patch based on the confidence scores. \emph{However we only update the labels of patches from diseased pavement images, since the labels of patches from the normal images should always be 0 ("normal").} IRAT is the core of the $E$ step.

\noindent\textbf{Image-based Rank Aware Threshold (IRAT):~} A patch $p_t$ from a diseased image $x_i$ labelled as the diseased patch by IRAT should meet any of the following two conditions:

(a) Its confidence score is above the ratio of the number of the diseased patch to the total one, $s_{j-1}$, in the previous iteration, and it can be automatically calculated in each iteration with the initialization, $s_0=0.5$;

(b) Its confidence score belongs to the top $r$ percentage high score in its image.

Such a label updating strategy can be mathematically denoted as follows:
\begin{equation}\label{updating}
\tilde{l}^i_{t(j)} =  \left\{
\begin{matrix}
\mathbf{{\rm 1}}, & {g^i_{t(j-1)}> \mathrm{min}(s_{j-1}, \mathcal{T}_i(r)) ~\mathrm{and}~ y_i =1}\\
\mathbf{{\rm 0}}, & {\mathrm{others}~~~~~~~~~~~~~~~~~~~~~~~~~~~~~~~~~~~~~~~}
\end{matrix}\right.
\end{equation}
where $\mathcal{T}_i(\cdot)$ returns the minimum threshold in the top $r$ percentage specific to the $i$-th image. In our implementation, $r=0.45$, which is empirically learned in a small size validation set.
\subsubsection{Prior Knowledge Biased Cross-Entropy}
\label{sssec:subsubhead}

We think that the labels of the diseased patches that own the higher confidence scores produced by PLIN in the previous iteration are more reliable than the ones who own the lower scores, and an improved PLIN should also suppress the normal patches that own high confidence scores. Therefore, we deem the confidence scores obtained and the distribution of the patch labels in the previous iteration as prior knowledge, and incorporate them to design a weighting scheme for cross-entropy. We introduce this novel cross-entropy loss named Prior Knowledge Biased Cross-Entropy (PKBCE) to the PLIN,
\begin{equation}\label{loss}\small
\begin{split}
\mathcal{L}_j &= -\frac{1}{nm}\sum_{i=1}^{n}\sum^{m}_{t=1}\frac{g^i_{t(j-1)}}{s_{j-1}}\{\tilde{l}^i_{t(j)}\mathrm{log}(g^i_{t(j)})\\
& +(1-\tilde{l}^i_{t(j)})\mathrm{log}(1-g^i_{t(j)})\}.
\end{split}
\end{equation}
The $\frac{g^i_{t(j-1)}}{s_{j-1}}$ is considered as the normalized version of $g^i_{t(j-1)}$, and a higher $g^i_{t(j-1)}$ implies that the corresponding patch is paid more attention to the next training.

\subsection{Pavement Disease Detection}

After the optimization of PLIN is converged, the trained PLIN model is used to label the patches of test images. And the detection label of a test image $x_i$ is achieved by the maximum pooling of its patch labels, $y_i=\mathrm{max}(\{\tilde{l}^i_t|\forall t\})$. {\em According to such a strategy, the final detection label inference is not up to the patch number of an image. In other words, our model can handle any resolution image.}

Algorithm~\ref{ioplin} presents the specific steps of our approach.
\begin{algorithm}
\caption{IOPLIN-based Pavement Disease Detection}
\label{ioplin}
\begin{algorithmic}[1]
\STATE {\bf \textit{[Initialization]}}
\STATE $\tilde{l}^i_{t(0)}=y_i \subset \{0,1\}$
\WHILE{$\sum_{i=1}^{n}\sum^{m}_{t=1}|\tilde{l}^i_{t(j-1)}-\tilde{l}^i_{t(j)}|>0$}
\STATE {\bf \textit{[M-step]}}
\STATE  Training ${M}_ \alpha \left ( \cdot  \right )$ with $\{x_i|\forall i\}$, $\{\tilde{l}^i_{t(j-1)}|\forall t,i\}$ and $\alpha_{j-1}$ using the loss in Equation~\ref{loss}
\STATE \noindent {\bf \textit{[E-step]}}
\STATE Obtaining the $g^i_{t(j)}$ for each patch and updating the labels of all patches based on Equation~\ref{updating} with $r$ and $s_{j-1}$.
\STATE Updating $s_{j}$ and the loss function in Equation~\ref{loss} based on new labels
\ENDWHILE
\STATE The detection label is the maximum pooling of the labels of its patches inferred by the trained PLIN.
\end{algorithmic}
\end{algorithm}

To speed up the convergence, the PLIN is also fine-tuned with thumbnails of the training pavement images before the iterative optimization. Our empirical study shows that such a trick is quite effective and can further improve the performance of IOPLIN. The details will be discussed in the experimental section.

\subsection{The Merits of IOPLIN}
In contrast to other deep learning models, IOPLIN enjoys many merits:
\begin{itemize}
\item IOPLIN is essentially a flexible local-based deep learning framework. Any CNN models can be plugged into IOPLIN as the backbone network.

\item IOPLIN can handle any resolution image and sufficiently exploit the image information. {\em If the image size is smaller than $300\times300$, IOPLIN will degenerate as a regular EfficientNet model.}

\item IOPLIN pays more attention to the local visual feature, and is able to roughly localize the diseased areas without using any patch-level prior supervised information.

\item IOPLIN significantly outperforms state-of-the-art CNN models, particularly in the high recall case.

\end{itemize}

\section{EXPERIMENTS AND RESULTS}
\label{sssec:majhead}

\subsection{Dataset and Setup}
\label{sssec:subhead}
\subsubsection{Dataset}
\begin{figure}[t]
\begin{minipage}[b]{1.0\linewidth}
  \centering
  \centerline{\includegraphics[width=7.5cm]{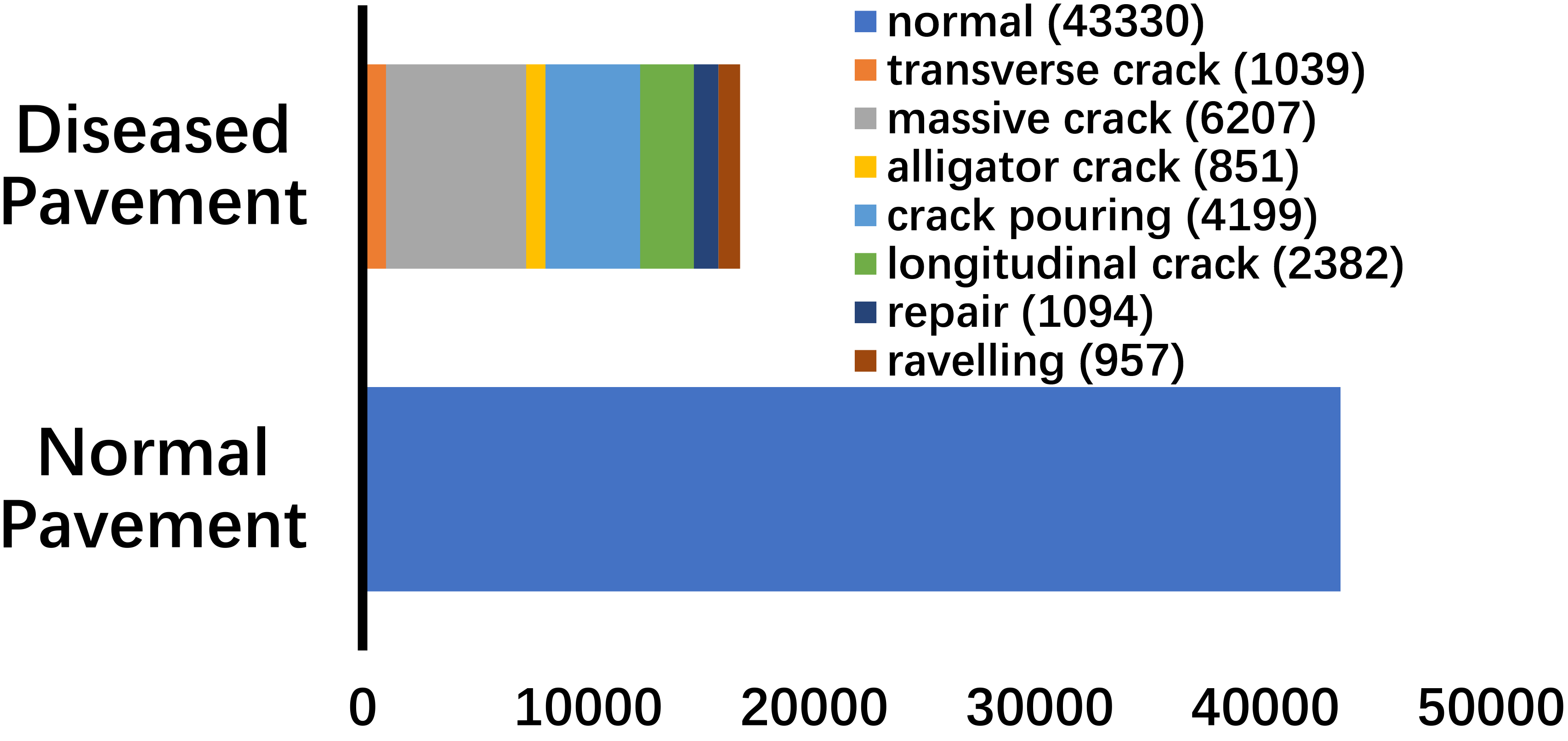}}
\end{minipage}
\vspace{-0.1cm}
\caption{The sample distribution of Chongqing University Bituminous Pavement Disease Detection (CQU-BPDD) dataset.}
\label{bpdd}
\vspace{-0.3cm}
\end{figure}

We release a novel large-scale Bituminous Pavement Disease Detection dataset named CQU-BPDD for evaluation. The CQU-BPDD dataset consists of 60,059 bituminous pavement images, which were automatically captured by the in-vehicle cameras of the professional pavement inspection vehicle at different times from different areas in southern China. Each pavement image is corresponding to a $2\times3$ meters pavement patch of highways and its resolution is 1200$\times$900. The CQU-BPDD dataset involves seven different distresses, namely \textit{transverse crack}, \textit{massive crack}, \textit{alligator crack}, \textit{crack pouring}, \textit{longitudinal crack}, \textit{raveling}, \textit{repair}, and the normal ones. The CQU-BPDD dataset is able to be accessed on {\url{https://dearcaat.github.io/CQU-BPDD}}. The data distribution of CQU-BPDD dataset is shown in Figure~\ref{bpdd}.

We randomly select 5,140 diseased pavement images involving all diseases and 5,000 normal pavement images to produce the training set, while the rest of dataset is used as the testing set. In the testing set, there are 11,589 diseased pavement images and 38,330 normal images.

\subsubsection{Evaluation Metrics}
Pavement disease detection is essentially a binary image classification task, so we adopt two well known binary classification performance evaluation metrics, namely precision and recall, for measuring performances. Let the diseased and normal samples be positives and negatives respectively. \textbf{P}recision is to measure how many the samples are the real positive samples (True Positives) among samples that are predicted as the positive samples by the learning model, while \textbf{R}ecall measures how many the real positive samples are correctly detected among all positive samples. The precision and recall can be mathematically represented as follow,
\begin{equation}\label{PR}
\textbf{P}=\frac{TP}{TP+FP},\quad\textbf{R}=\frac{TP}{TP+FN},\nonumber
\end{equation}
where $TP$, $FP$ and $FN$ are the numbers of true positives, false positives and false negatives respectively. However, the precision and recall are changeable via adjusting the threshold of confidence score (or classification boundary). More specifically, a binary classification system often assigns a confidence score to each image, and such confidence scores indicate the probabilities of images belong to the positive categories. If the confidence score of an image is higher than the threshold, the image will be classified as a positive sample. In such a manner, a higher threshold often leads to the higher precision but the lower recall, and vice versa. In the medical or pavement image analysis tasks, it is more meaningful to discuss the precision under the high recall, since the miss of the positive sample (the diseased sample) will lead to more serious impact than the miss of the negative one.

Additionally, we adopt the Area Under Curve (\textbf{AUC}) of Receiver Operating Characteristic (ROC)~\cite{auc,roc} as the comprehensive performance evaluation metric, whose value is independent from the setting of the threshold. The AUC can well reflect the potential performance of a binary classifier. It is mathematically defined as follows,
\begin{equation}\label{AUC}
AUC=\frac{S_{p}-N_{p}\left ( N_{p}+1 \right )/2}{N_{p}N_{n}}
\end{equation}
where $S_{p}$ is the sum of the all positive samples ranked, while $N_{p}$ and $N_{n}$ denote the number of positive and negative samples. More details about precision, recall and AUC can refer to~\cite{auc,roc}.

\subsubsection{Compared Methods}
 Seven well-known image classification approaches, namely, Histogram of Oriented Gradient (HOG)~\cite{hog}, Local Binary Pattern (LBP)~\cite{lbp}, Fisher Vector (FV)~\cite{fisher,fv}, VGG-19~\cite{simonyan2014}, ResNet-50~\cite{he2016}, Inception-v3~\cite{szegedy2016}, EfficientNet-B3~\cite{tan2019} are used for comparison. HOG and LBP are the local-based hand craft representation methods while FV is a shallow learning-based representation methods. The last four ones are the state-of-the-art deep learning approaches that have a similar number of parameters and have been successfully applied to numerous image classification tasks.
 All the hyper-parameters involved in the compared methods are well-tuned.

\begin{table}[t]
\caption{The comparison of different automatic pavement disease detection models. (P@R = $n$\% indicates the precision when the corresponding recall is equal to $n$\%, PCA = Principal Component Analysis, SVM = Support Vector Machine )}
\centering
\vspace{-0.3cm}
\begin{tabularx}{8cm}{lX<{\centering}X<{\centering}X<{\centering}}
\hline
Method & AUC & P@R=90\% & P@R=95\% \\
\hline
HOG+PCA+SVM~\cite{hog}& 77.7\% &31.2\%&28.4\%\\
LBP+PCA+SVM~\cite{lbp}&82.4\% &34.9\%&30.3\%\\
HOG+FV+SVM~\cite{fv}&88.8\% &43.9\%&35.4\%\\
ResNet-50~\cite{he2016}  & 90.5\% & 45.0\% & 35.3\%\\
Inception-v3~\cite{szegedy2016} & 93.3\% & 56.0\% & 42.3\% \\
VGG-19~\cite{simonyan2014}  & 94.2\% & 60.0\% & 45.0\%\\
EfficientNet-B3~\cite{tan2019} & 95.4\% & 68.9\% & 51.1\%\\
{\bf IOPLIN (Ours)} & {\bf 97.4\%} & {\bf 81.7\%} & {\bf 67.0\%}\\
\hline
\end{tabularx}
\label{compare}
\vspace{-0.1cm}
\end{table}

\begin{figure}[t]
\begin{minipage}[b]{1.0\linewidth}
  \centering
  \centerline{\includegraphics[width=7cm]{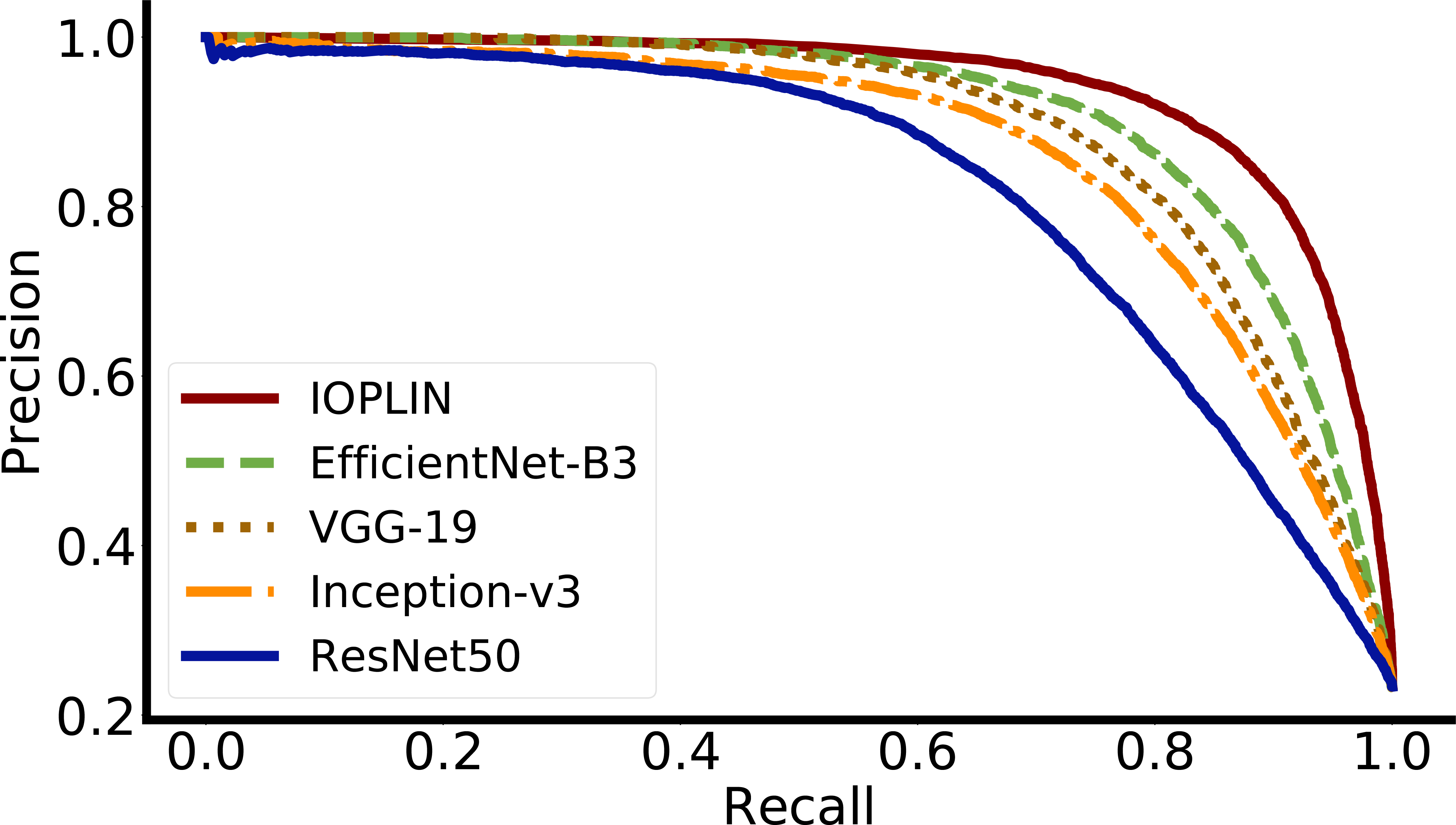}}
\end{minipage}
\vspace{-0.7cm}
\caption{Precision-Recall (P-R) curves of different deep learning models.}
\label{pr}
\vspace{-0.5cm}
\end{figure}

\begin{figure}[t]
\begin{minipage}[b]{1.0\linewidth}
  \centering
  \centerline{\includegraphics[width=7cm]{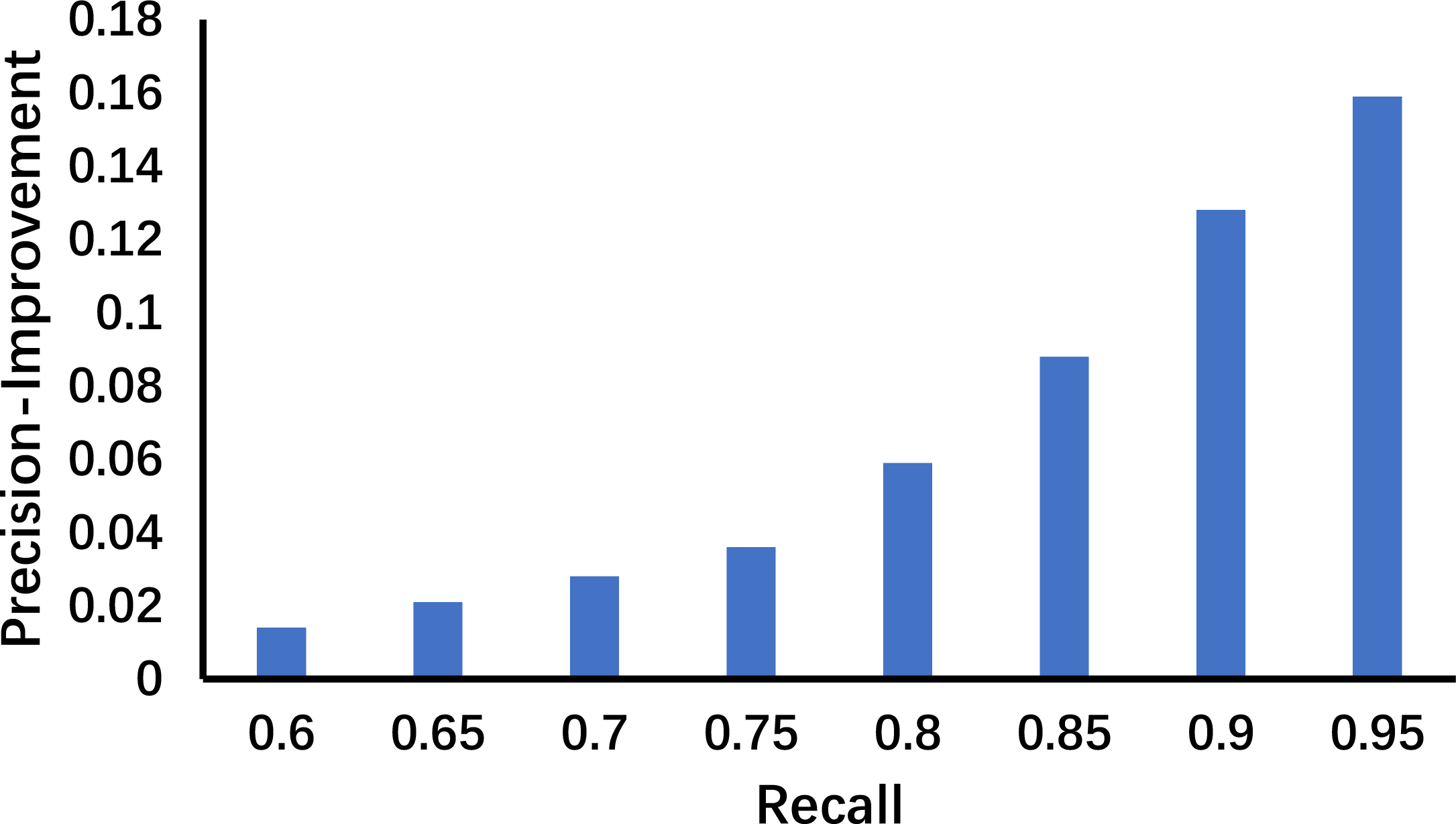}}
\end{minipage}
\vspace{-0.5cm}
\caption{The precision improvements of IOPLIN over EfficientNet-B3, which is the best performed compared approach, in different recalls. }
\label{improves}
\vspace{-0.4cm}
\end{figure}

\subsection{Pavement Disease Detection}
Table~\ref{compare} tabulates the detection performances of different deep learning approaches and Figure~\ref{pr} shows the P-R curves of these methods. From these results, it is clear that our work consistently outperforms the compared methods with a significant advantage in different evaluation metrics. EfficientNet-B3 achieves the best performance among the seven compared methods, and it is also adopted as the backbone of our Patch Label Inference Network (PLIN). Even so, our work gets 2\% gains in AUC over EfficientNet-B3, and the precision gains of our work over it are 12.8\% and 15.9\% when the recalls are fixed to 90\% and 95\% respectively. The hand-craft feature or shallow learning-based methods perform much worse than the deep learning ones obviously. They even cannot achieve 90\% accuracy in AUC. The performance gain in AUC of our method over the best performed handcraft feature-based approach, HOG + FV + SVM, is 8.6\%, and the precisions of our method are almost the double  the ones of HOG + FV + SVM when the recalls are fixed to 90\% and 95\%.

Figure~\ref{improves} shows the precision improvements of IOPLIN over EfficientNet-B3 in different recalls, which reveals an interesting phenomenon: the precision gain of our work over EfficientNet-B3 is increased along with the increase of recall. This is a very desirable property for automatic pavement disease detection, since people always pay more attention to the disease images rather than the normal ones. This is because omitting the disease images may cause serious safety risks, whereas omitting the normal ones almost leads to no cost in real life. In such a manner, a good pavement disease detection approach should perform much better in a higher recall. All results imply that our work is better to meet such requirements.

\subsection{Cross-Dataset Validation}
\begin{figure}[t]
\centering
\subfigure[Diseased image on CFD. ]{
\centering
\includegraphics[width=4.1cm]{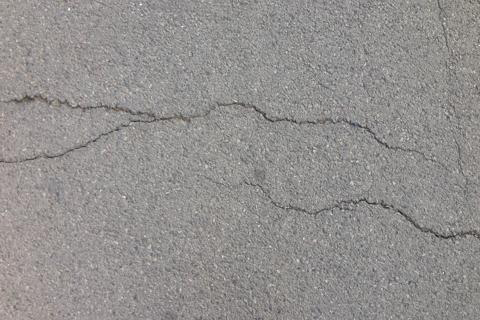}
\label{}}
\subfigure[Normal image on CFD]{
\centering
\includegraphics[width=4.1cm]{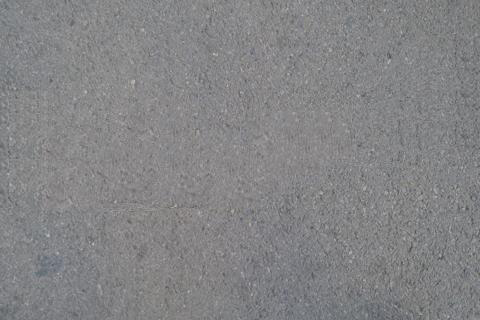}
\label{}}
\subfigure[Diseased image on CrackTree200. ]{
\centering
\includegraphics[width=4.1cm]{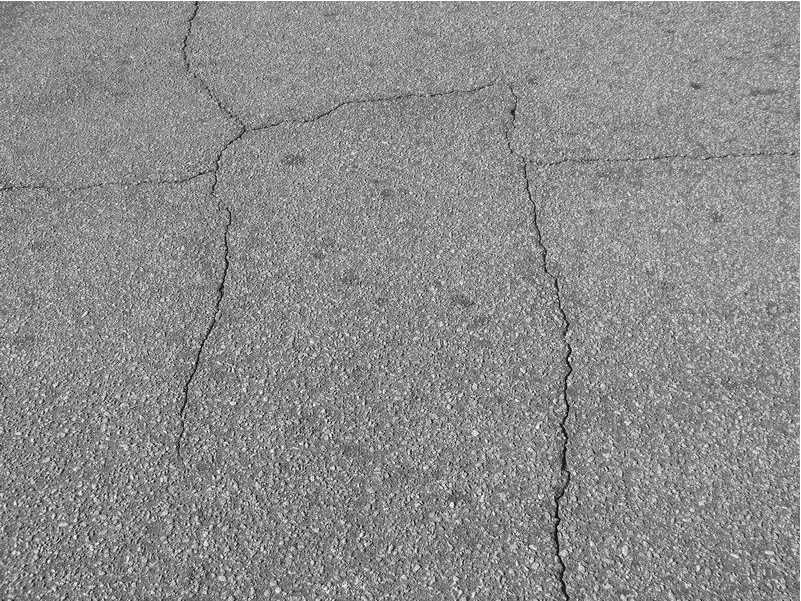}
\label{}}
\subfigure[Normal image on CrackTree200.]{
\centering
\includegraphics[width=4.1cm]{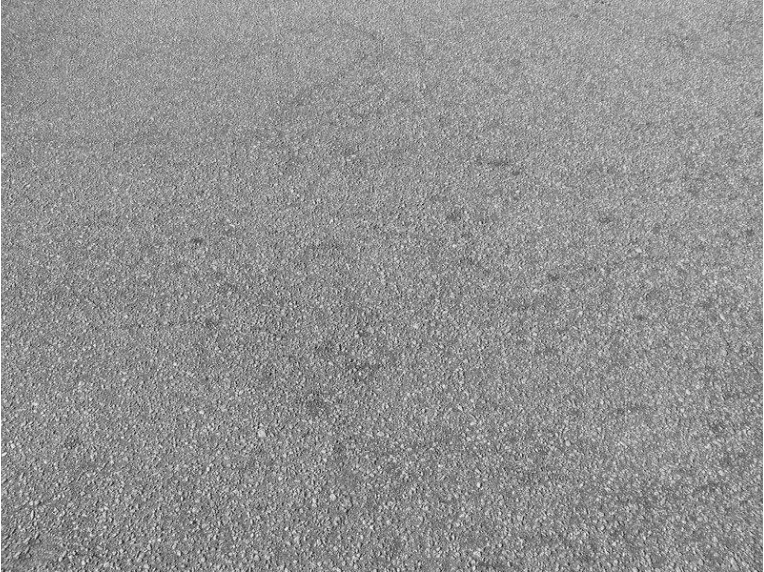}
\label{}}
\vspace{-0.3cm}
\caption{ The pavement image examples on CFD and cracktree200 datasets. Note, the normal ones are recovered from their diseased version via replacing the diseased pixels with their neighbor pixels based on the pixel-level labels provided by the datasets. }\label{cross-case}
\vspace{-0.4cm}
\end{figure}

In order to validate the generalization ability of our method to other data, two commonly used pavement crack segmentation (pixel-level pavement crack detection) datasets, namely Crack Forest Dataset(CFD)~\cite{shi2016} and CrackTree200~\cite{cracktree}, are adopted for validation. CFD originally consists of 155 images whose resolutions are $480\times320$. CrackTree200 contains 206 images whose resolutions are $800\times600$. Since these two datasets are designed for studying the so-called road crack detection task which is essentially the pavement crack segmentation task in the perspective of computer vision, all samples on these two datasets are actually the diseased images (positive samples). In such a case, it is impossible to validate our pavement disease detection task due to the absence of the negative samples (the normal images). Therefore, we automatically produce the normal version of each diseased images via replacing the disease pixels with their neighbor normal pixels. However, such an automatic normal image fashion does not always work well for all samples. We manually filter out some low-quality generated normal images and only retain the high-quality ones. Finally, we have 155 diseased images and 114 recovered normal images on CFD dataset, while 206 diseased images and 191 recovered normal images on CrackTree200 dataset. Figure~\ref{cross-case} shows some examples of these two datasets. {\em Note, the images on these datasets are all only used for testing, and the involved pavement disease detection models in this section are only trained on CQU-BPDD dataset while without any fine-tuning on these two datasets.} The reason is that we intend to compare the cross-dataset generalization abilities between IOPLIN and EfficientNet-B3.

\begin{table}[t]
  \centering
  \caption{The performance comparison between our method and the baseline on other two pavement disease datasets}
  \label{cross-val}
  \vspace{-0.3cm}
  \begin{tabular}{c | c  p{0.7cm}<{\centering} p{1.1cm}<{\centering} p{1.1cm}<{\centering}}
  \hline
  Validation Dataset& Methods &AUC & P@R=90\% & P@R=95\% \\
  \hline
\multirow{2}{*}{CFD~\cite{shi2016}}&  EfficientNet-B3 & 95.1\% & 92.1\%&83.5\% \\
&\textbf{IOPLIN}&\textbf{96.0\%}&\textbf{95.9\%}&\textbf{91.3\%}\\
\hline
\hline
\multirow{2}{*}{CrackTree200~\cite{cracktree}}&  EfficientNet-B3 &98.0\%&94.8\%&94.6\%\\ 
&\textbf{IOPLIN}&\textbf{99.5\%}&\textbf{99.5\%}&\textbf{98.5\%}\\
\hline
  \end{tabular}
  \vspace{-0.5cm}
  \end{table}

Table~\ref{cross-val} shows the performances of IOPLIN and EfficientNet-B3 on CFD and CrackTree200 datasets. The experimental results confirm that IOPLIN still enjoys better performances than the EfficinetNet-B3 under all three evaluation indicators. More specifically, the precision gains of IOPLIN over EfficinetNet-B3 are 3.8\% and 7.8\% on CFD datasets when the recalls are fixed to 90\% and 95\% respectively, while these numbers on the CrackTree200 dataset are 4.7\% and 3.9\%. Moreover, even EfficientNet-B3 is already at a very high level of AUC on both of the two datasets. Our method still improves it. IOPLIN achieves 0.9\% and 1.5\% higher AUCs than EfficientNet-B3 on CFD and CrackTree200 datasets respectively. Clearly, all these observations imply that our method posses the better cross-data generalization ability in comparison with EfficientNet-B3 .

\subsection{Robustness Analysis}
\begin{figure}[t]
    \centering
    \includegraphics[scale=0.24]{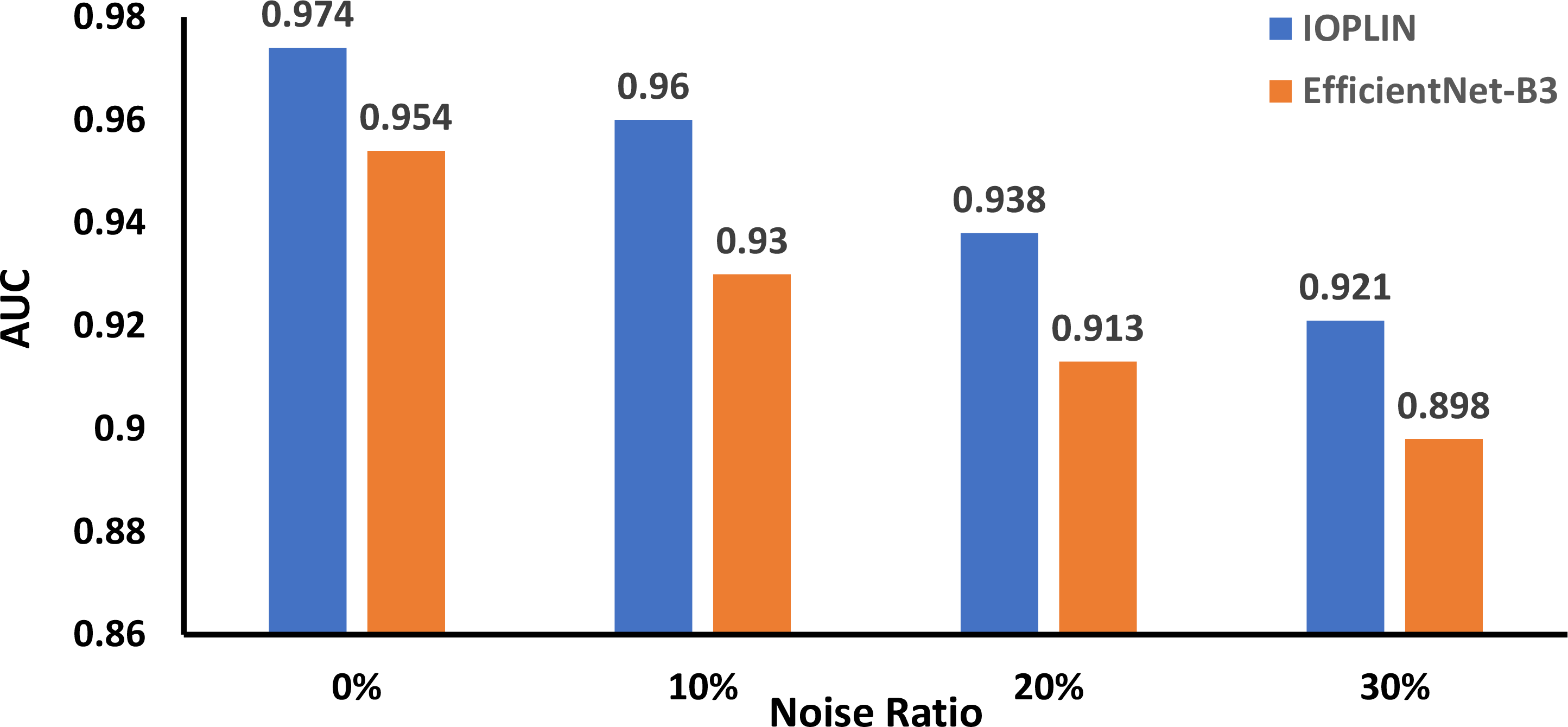}
    \caption{The performances of IOPLIN and EfficientNet B3 under different noise ratios in AUC.}
    \vspace{-0.1cm}
    \label{robustness}
\end{figure}

In this section, we conduct the robustness analysis of our method on CQU-BPDD dataset. The Gaussian noise is randomly introduced to each testing image to corrupt a proportion of pixels.  The noise ratio represents how many proportions of pixels in an image that have been corrupted. Figure~\ref{robustness} shows the performances of our method and EfficinetNet-B3 under different noise ratios. From observations, it is clear that our method consistently outperforms EfficientNet-B3 under all noise ratios. Moreover, the AUC gains of our method over EfficientNet-B3 are 2.0\% in the clean case, while these gains are 3.0\%, 2.7\% and 2.3\% under 10\%, 20\% and 30\% noise ratios respectively, which shows the stronger advantage of our method even in the noisy scenario. All these phenomena reveal that our method enjoys the stronger robustness to noise in comparison with its baseline, EfficientNet-B3.

\subsection{Ablation Study}
\begin{figure}[t]
\begin{minipage}[b]{1\linewidth}
  \centering
  \centerline{\includegraphics[width=8cm]{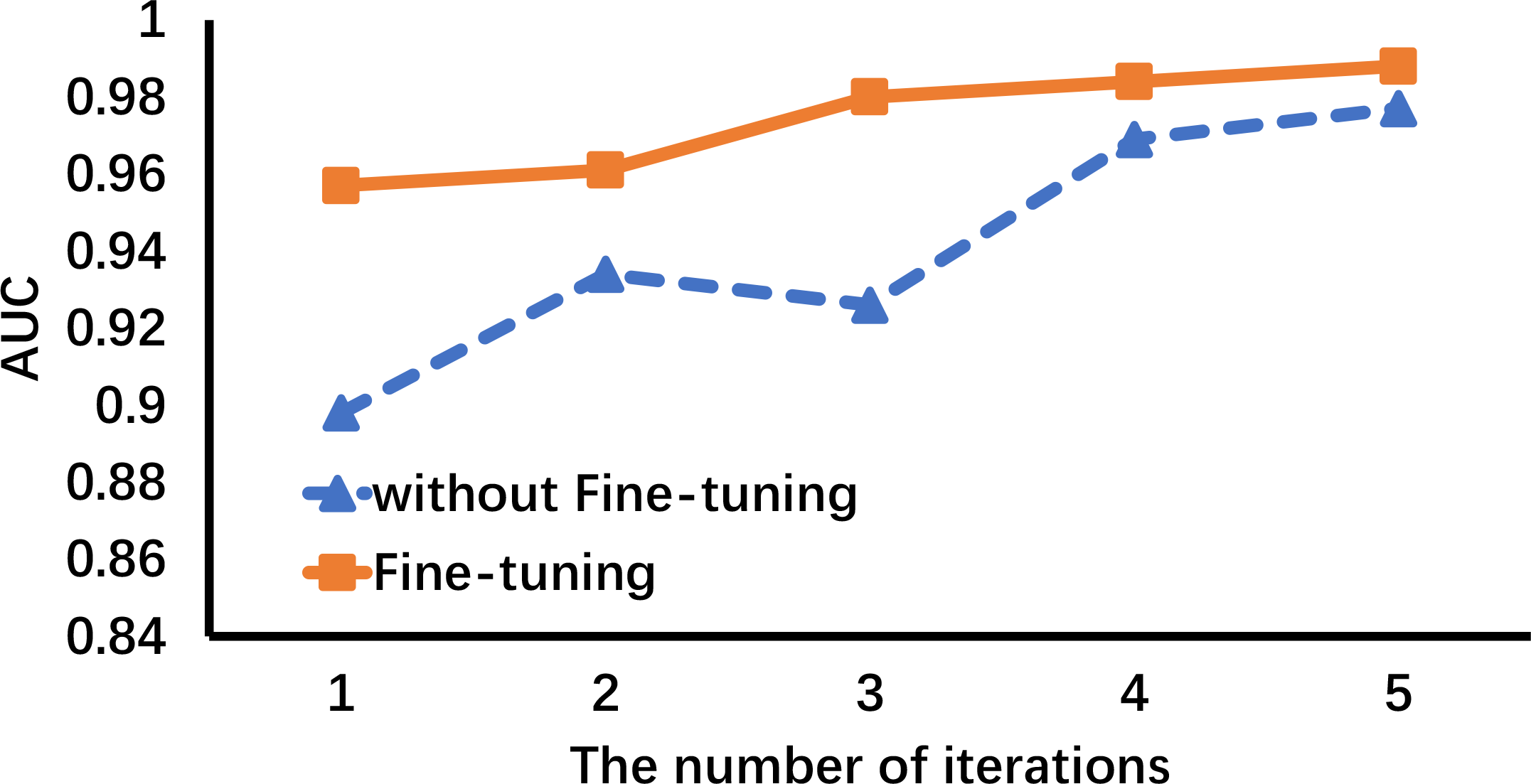}}
\end{minipage}
\vspace{-0.3cm}
\caption{The convergence analysis of IOPLIN with or without the thumbnails-based fine-tuning.}
\label{finetune}
\vspace{-0.1cm}
\end{figure}

Table~\ref{ablation} shows the ablation analysis results, where CLAHE, IRAT, FT and PKBCE, respectively, represent the contrast limited adaptive histogram equalization, image-based rank award threshold, fine-tuned with the thumbnails of the pavement images and prior knowledge biased cross entropy. The comparison of the first two rows implies that the CALHE step can slightly improve the pavement disease detection performances. The backbone network of PLIN is EfficientNet-B3. However, IRAT + CLAHE + PLIN performs slightly worse than CLAHE + EfficientNet-B3. We attribute this to the different training schemes of EfficientNet-B3 in these two approaches. The first one is iteratively trained for patch label inference without any patch label ground truth. In contrast, the latter one is adequately trained with the certain ground truths for image label inference. By considering the fine-tuning with the thumbnails, the IOPLIN gets 0.7\% AUC gain. This indicates that a good initialization of PLIN is helpful for optimizing the models. Among all the tricks in IOPLIN, PKBCE contributes the most, which improves IOPLIN 1\% in AUC. We also plot the relationship between the iteration number and the detection performance in Figure~\ref{finetune}. This reveals another benefit from fine-tuning which speeds up the convergence of the model optimization.

\begin{table}[t]
\vspace{-0.1cm}
\caption{The ablation analysis of IOPLIN.}
\centering
\vspace{-0.2cm}
\begin{tabularx}{8cm}{lX<{\centering}X<{\centering}X<{\centering}}
\hline
Method & AUC \\
\hline
EfficientNet-B3 (Baseline) & 95.4\% \\
CLAHE+EfficientNet-B3  & 95.9\%  \\
IRAT+CLAHE+PLIN  & 95.7\%  \\
FT+IRAT+CLAHE+PLIN &  96.4\% \\
{\bf PKBCE+FT+IRAT+CLAHE+PLIN (IOPLIN)} &  {\bf 97.4}\% \\
\hline
\end{tabularx}
\label{ablation}
\vspace{-0.3 cm}
\end{table}

Moreover, we also empirically discuss the effects of different Histogram Equalization (HE) on the performances of our method. 
Table~\ref{pre-processing} tabulates the performances of our method using different pre-processed pavement images.  From observations, our adopted pre-processing technique, named Contrast Limited Adaptive Histogram Equalization (CLAHE)~\cite{pizer1990}, performs the best under all three evaluation indicators. More specifically, CLAHE gets 0.9\% and 1.5\% more gains in AUC compared with no processing and regular HE respectively. This reveals two facts that CLAHE can benefit the automatic pavement disease detection system while an improper pre-processing, such as regular HE, may lead to performance degradation.


\begin{table}[t]
\vspace{-0.1cm}
\caption{The effects of pre-processing to the performance of our method.}
\centering
\vspace{-0.2cm}
\begin{tabularx}{8cm}{lX<{\centering}X<{\centering}X<{\centering}}
\hline
Method &  AUC & P@R=90\% & P@R=95\%\\
\hline
Original Image&96.5\% & 74.9\%& 60.1\%\\
Regular HE& 95.9\%&71.2\%&57.1\% \\
CLAHE   & \textbf{97.4\%}&\textbf{81.7\%}&\textbf{67.4\%}  \\
\hline
\end{tabularx}
\label{pre-processing}
\vspace{-0.3cm}
\end{table}

\subsection{User Scenarios}
Our developed automatic pavement disease detection technique can possibly be used in many scenarios. The pavement image screening and patch-level pavement disease localization are two typical applications of our method.

\begin{figure}[t]
\centering
\subfigure[The example of normal image filtering out when threshold = 0.5.]{
\centering
\includegraphics[width=8cm]{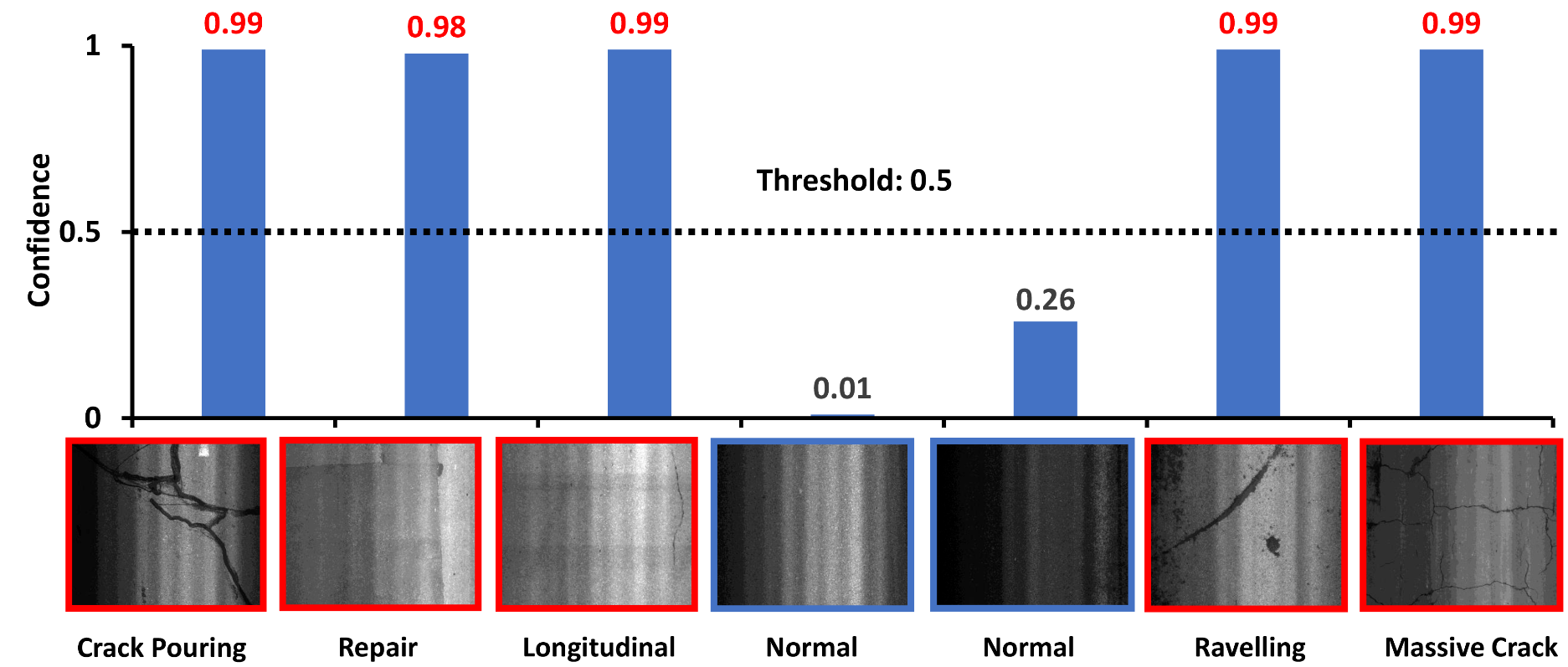}
\label{}}
\subfigure[The example of normal image filtering out when threshold = 0.8.]{
\centering
\includegraphics[width=8cm]{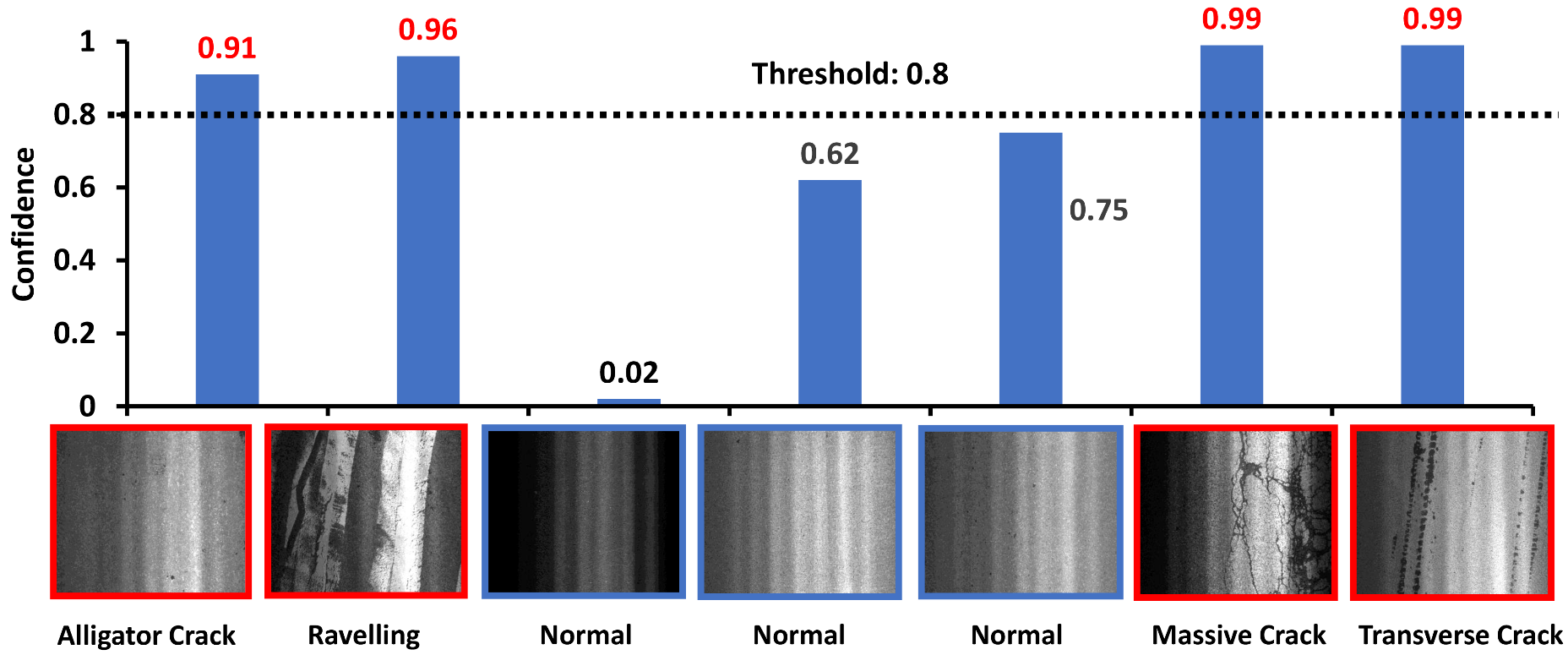}
\label{}}
\caption{ The examples of using different thresholds to filter out the normal pavement images in two randomly sampled batches. The red ones indicate their confidence scores are above the threshold, and considered as the diseased samples. The bottom is the real categories of these images. }\label{filter}
\end{figure}

In our pavement disease detection system, it produces a confidence score in the range of 0 to 1 for each pavement image to measure its probability that the image belongs to the diseased one. In such a manner, we can filter out most of normal pavement images by setting a threshold of confidence score. Then the engineers only need to manually analyze a few of pavement images which clearly reduces the costs of both labor and time. Here, we give two examples for showing how to use our method for pavement disease screening in Figure~\ref{filter}. More specifically, we input two query pavement image batches randomly sampled from the testing set, and two different thresholds are used for filtering out the normal samples for these two batches, which all consist of seven pavement images. In these examples, we can adopt a proper threshold for filtering out all the normal pavements, and preserve the diseased ones for further analysis. In our testing set (49,919 samples), 75\% of normal pavement images can be filtered out and 97.4\% of diseased pavement images are correctly detected when the threshold is set to 0.5. If we set the threshold to 0.8, more than 90\% of normal images can be filtered out and 93.3\% of diseased pavement images are correctly detected. The threshold is essentially the classification boundary, which is tunable in real applications. A higher threshold means the stricter disease criterion, which leads to filtering out more normal samples but more diseased samples are falsely labelled. In the pavement disease analysis, we suggest setting a lower threshold to avoid the misclassification of too many diseased samples.

\begin{figure}[t]
\centering
\subfigure[ ]{
\centering
\includegraphics[width=4cm]{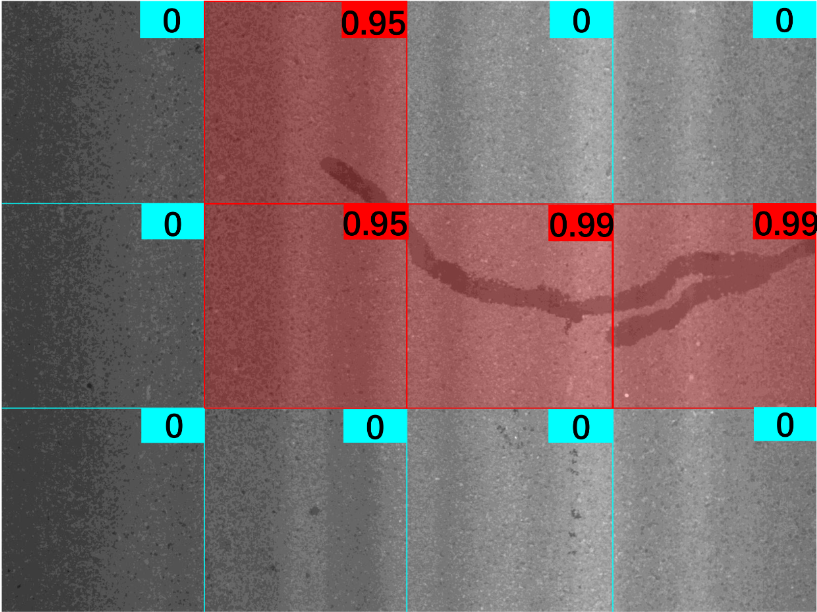}
\label{}}
\subfigure[]{
\centering
\includegraphics[width=4cm]{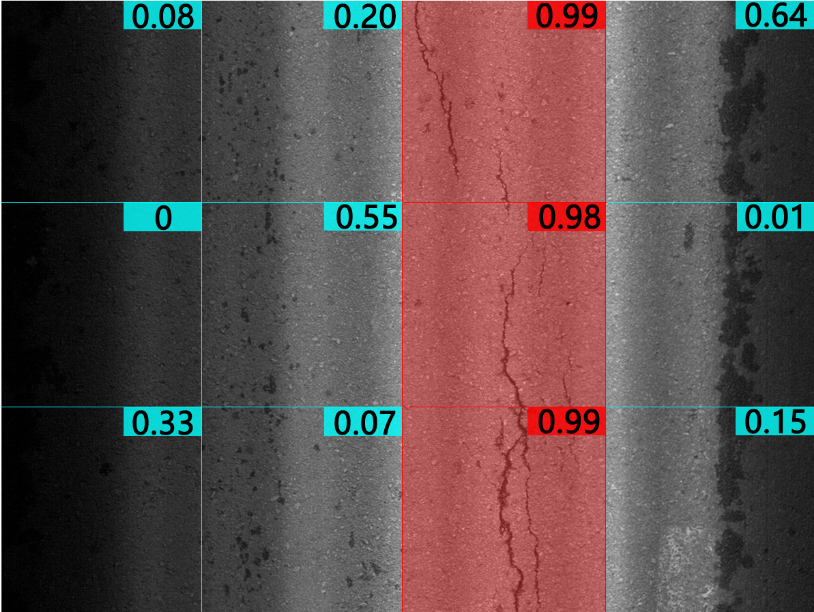}
\label{}}
\caption{ The visualizations of the several pavement images in patch-level, and the red patches mean the diseased ones. Our method provides a weakly supervised way for localizing the diseased areas in the image patch level. }\label{patch}
\end{figure}

%
Different from the conventional detection regime, IOPLIN accomplishes detection by judging if there exists any diseased patch in the image. In such a strategy, the labels of patches in an image can be roughly inferred, and these labels contain important by-product information that explains and even benefits the solution of the following-up tasks. We visualize the inferred labels with confidence scores of patches from two testing images in Figure~\ref{patch}. The observations show that the patch labels inferred by our method can further localize the diseased areas in patch level without any prior location information for training.

\section{Conclusion}
\label{con}
In this paper, we proposed a novel deep learning framework named Iteratively Optimized Patch Label Inference Network (IOPLIN) for automatic pavement disease detection. IOPLIN iteratively trains the Patch Label Inference Network (PLIN) only with the image labels by applying the EM Inspired Patch Label Distillation strategy. Then it infers the patch labels for a testing pavement image and accomplishes the detection task by maximum pooling of its patch labels. A novel large-scale Bituminous Pavement Disease Detection dataset named CQU-BPDD was constructed for evaluating the effectiveness of our work. The experimental results demonstrate the superiority of our method in comparison with state-of-the-art CNN approaches and also show that IOPLIN is capable of localizing the diseased areas without any prior information about the location.


%

%

\section*{Acknowledgment}
The work described in this paper was partially supported by National Natural Science Foundation of China (No. 61602068), Fundamental Research Funds for the Central Universities (No. 106112015CDJRC091101) and the Science and Technology Research Program of Chongqing Municipal Education Commission of China under Grant No. KJQN201800705 and KJQN201900726.

%
%



%

\bibliographystyle{IEEEtran}
\bibliography{IEEEabrv,IEEEexample}
%
\vspace{-2cm}
\begin{IEEEbiography}[{\includegraphics[width=1in,height=1.25in,clip,keepaspectratio]{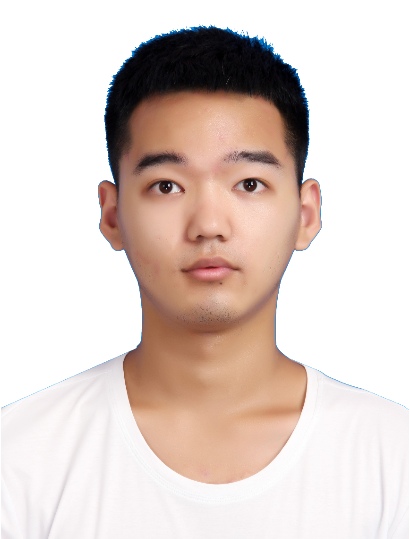}}]{Wenhao Tang} is currently a research Intern at Ministry of Education Key Laboratory of Dependable Service Computing in Cyber Physical Society, Chongqing University, Chongqing, P.R. China. He will earn his bachelor in 2021 and then continue to pursue his master degree in software engineering at Chongqing University. His research interests include intelligent transportation systems, computer vision and image processing.
\end{IEEEbiography}
\vspace{-1.9 cm}
\begin{IEEEbiography}[{\includegraphics[width=1in,height=1.25in,clip,keepaspectratio]{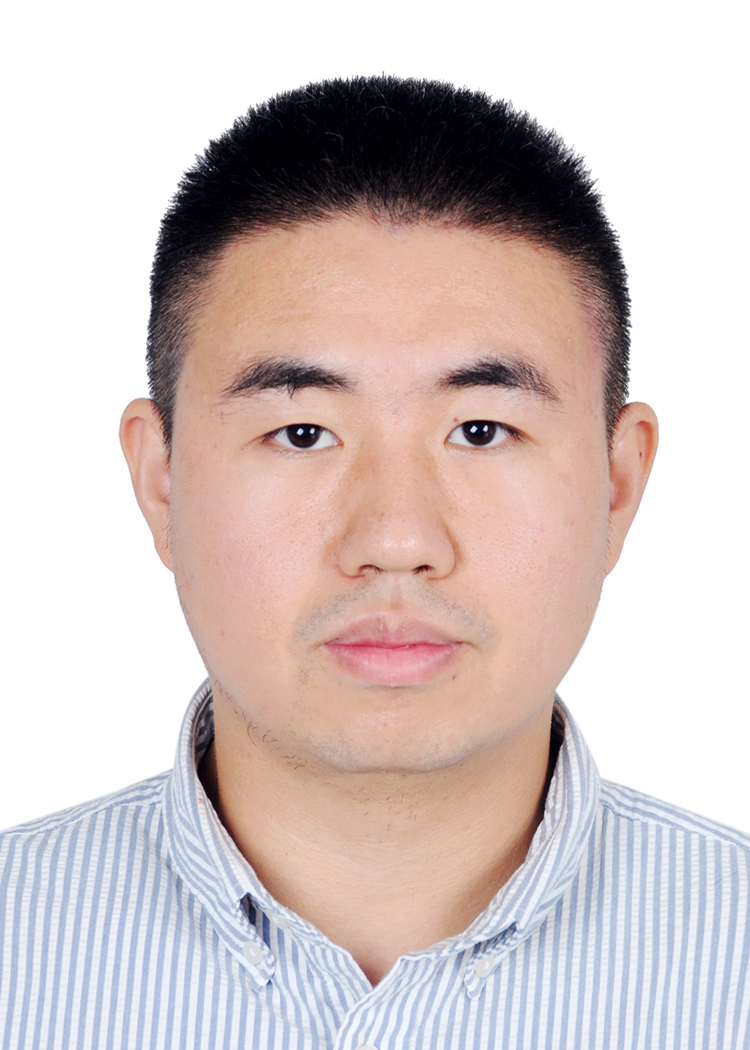}}]{Sheng Huang} (M'15)
 received his BEng and PhD degrees both from Chongqing University, Chongqing, P.R.China, in 2010 and 2015 respectively. He was also a visiting PhD student at the department of computer science, Rutgers University, New Brunswick, NJ, USA, from 2012 to 2014. He is currently an associate professor at the school of big data and software engineering, Chongqing University, and also affiliated with the Ministry of Education Key Laboratory of Dependable Service Computing in Cyber Physical Society. He has authored/coauthored more than 40 scientific papers in venues, such as CVPR, AAAI, TIP, TIFS, TMI and TCSVT. His research interests include computer vision, machine learning, image processing and artificial intelligent applications.
\end{IEEEbiography}
\vspace{-1.9cm}
\begin{IEEEbiography}[{\includegraphics[width=1in,height=1.25in,clip,keepaspectratio]{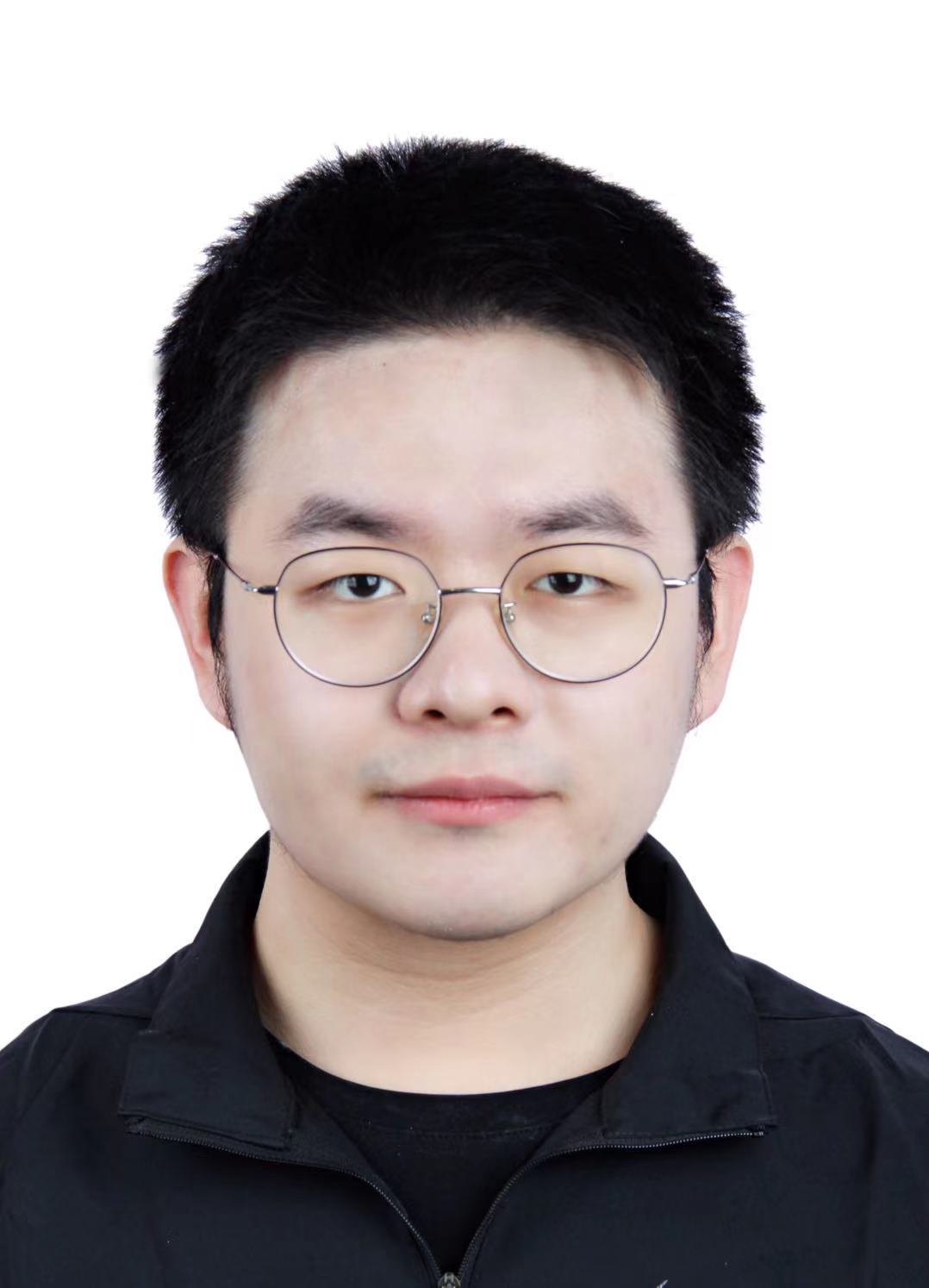}}]{Qiming Zhao} is currently a research Intern at Ministry of Education Key Laboratory of Dependable Service Computing in Cyber Physical Society, Chongqing University, Chongqing, P.R. China. He will earn his bachelor in 2021 at Chongqing University and pursue his master degree at University of Pittsburgh, USA. His research interests include computer vision and data mining.
\end{IEEEbiography}
\vspace{-10 mm}
\begin{IEEEbiography}[{\includegraphics[width=1in,height=1.25in,clip,keepaspectratio]{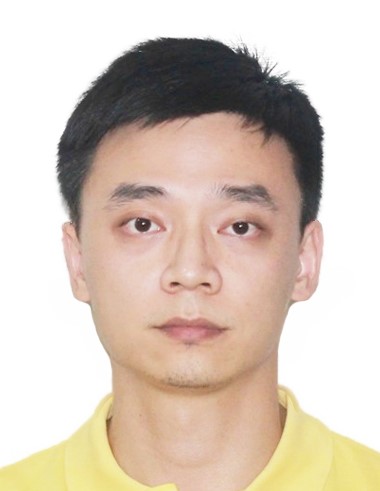}}]{Ren Li} received the Ph.D. degree in computer science from Chongqing University, Chongqing, China, in 2013.  He is currently an associate professor at the School of Information Science and Engineering, Chongqing Jiaotong University, Chongqing, China. His research interests include knowledge graph and structural health monitoring technologies.
\end{IEEEbiography}
\vspace{-17cm}
\begin{IEEEbiography}[{\includegraphics[width=1in,height=1.25in,clip,keepaspectratio]{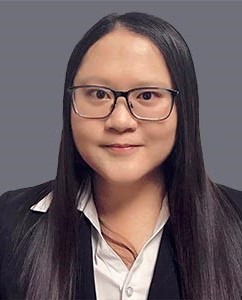}}]{Luwen Huangfu}
received her Ph.D. degree in Management Information Systems from the University of Arizona (UA), Arizona, USA. She obtained M.S. degree in Computer Science from Chinese Academy of Sciences (CAS), Beijing, and B.S. degree in Software Engineering from Chongqing University (CQU), Chongqing, P.R. China. She is currently an Assistant Professor at Fowler College of Business, San Diego State University (SDSU), California, USA. She has authored/coauthored more than 20 scientific papers in venues, such as AMCIS, LREC, PAJAIS, ISI, and ICME. Her research interests include business analytics, text mining, data mining, artificial intelligence and healthcare management.
\end{IEEEbiography}


%
%




\end{document}